\documentclass{article}
\usepackage{ijcai23}

\usepackage{times}
\usepackage{soul}
\usepackage{url}
\usepackage[hidelinks]{hyperref}
\usepackage[utf8]{inputenc}
\usepackage[small]{caption}
\usepackage{graphicx}
\usepackage{amsmath}
\usepackage{amsthm}
\usepackage{booktabs}
\usepackage{algorithm}
\usepackage{algorithmic}
\usepackage[switch]{lineno}
\usepackage{xcolor}
\usepackage{multirow}
\usepackage{url}
\usepackage{hyperref}

\title{Refining the Optimization Target for Automatic Univariate Time Series Anomaly Detection in Monitoring Services}

\author{
    Manqing Dong\and Zhanxiang Zhao \and Yitong Geng \and Wentao Li \and Wei Wang \And Huai Jiang
    \affiliations
    eBay Inc.
    \emails
    \{madong, zhanzhao, yigeng, wenli, wwang18, huajiang\}@ebay.com
}

\begin{document}

\maketitle

\begin{abstract}
Time series anomaly detection is crucial for industrial monitoring services that handle a large volume of data, aiming to ensure reliability and optimize system performance. Existing methods often require extensive labeled resources and manual parameter selection, highlighting the need for automation. This paper proposes a comprehensive framework for automatic parameter optimization in time series anomaly detection models. The framework introduces three optimization targets: prediction score, shape score, and sensitivity score, which can be easily adapted to different model backbones without prior knowledge or manual labeling efforts. The proposed framework has been successfully applied online for over six months, serving more than 50,000 time series every minute. It simplifies the user's experience by requiring only an expected sensitive value, offering a user-friendly interface, and achieving desired detection results. Extensive evaluations conducted on public datasets and comparison with other methods further confirm the effectiveness of the proposed framework.
\end{abstract}

\section{Introduction}
Industrial monitoring services typically oversee millions of time series data points on a daily basis, making timely and accurate time series anomaly detection vital for maintaining reliability and optimizing the performance of diverse systems and applications.

Various methods have been employed in the field of time series anomaly detection. For instance, statistical models analyze patterns in time series data using a historical time window and identify time points with extreme deviations as anomalies~\cite{breunig2000lof,siffer2017anomaly}. Forecasting-based approaches, including traditional models like moving average~\cite{yu2016improved}, as well as neural sequence models such as LSTM~\cite{hochreiter1997long} and Transformer~\cite{zhou2021informer}, aim to predict values based on a given range of time series inputs, with anomalies characterized by significant deviations from the predicted values. On the other hand, reconstruction-based methods concentrate on reconstructing time series data, assuming that anomalies exhibit high reconstruction errors~\cite{sabokrou2015real}.

Despite the effectiveness of existing time series anomaly detection methods, they often require extensive labeled resources to achieve optimal performance. Additionally, manual and careful parameter selection is necessary to accommodate diverse needs. This underscores the urgent need for automation in time series anomaly detection, as it can alleviate the reliance on labeled resources and automate the parameter selection process.

Existing strategies for automatic parameter tuning can be classified into three categories. The first approach, as demonstrated in Prophet \cite{taylor2018forecasting}, involves optimizing parameters based solely on prediction errors. However, relying solely on prediction metrics can result in overfitting to anomalies, leading to increased false negatives.
The second approach treats parameter tuning as a prediction task itself, where a model is trained to directly predict the best parameters for a given algorithm \cite{zhang2021self,chatterjee2022mospat}. This method requires prior knowledge and a significant amount of labeled resources for each algorithm. Additionally, when a new model is introduced, manual parameter tuning is necessary again to generate the correct parameter labels.
The third approach treats anomaly detection as a binary classification problem and optimizes parameters based on anomaly labels \cite{lai2021tods}. However, this approach faces challenges when applied to industrial monitoring platforms. It heavily relies on labeled anomalies in the training data, which are often missing or difficult to obtain in general monitoring services. Moreover, this approach is not suitable for handling new incoming time series data, rendering the model unavailable for real-time applications.

The aforementioned automatic parameter tuning solutions are not applicable to monitoring services due to the following reasons.
First, monitoring services typically handle an immense volume of time series data, making it impractical to label anomaly points for each individual time series.
Second, monitoring services cater to diverse user groups, each with varying needs and sensitivities towards anomalies. For instance, some users may prefer an anomaly detector that only reports the most severe anomalies, while others may want to monitor all potential anomalies.
Furthermore, users often lack in-depth expertise in anomaly detection algorithms. If the optimized parameters do not meet their requirements, users would need to invest significant time and effort to gain proficiency in the detection algorithms for further fine-tuning.

\begin{figure}[t]
    \centering
    \includegraphics[width=0.8\linewidth]{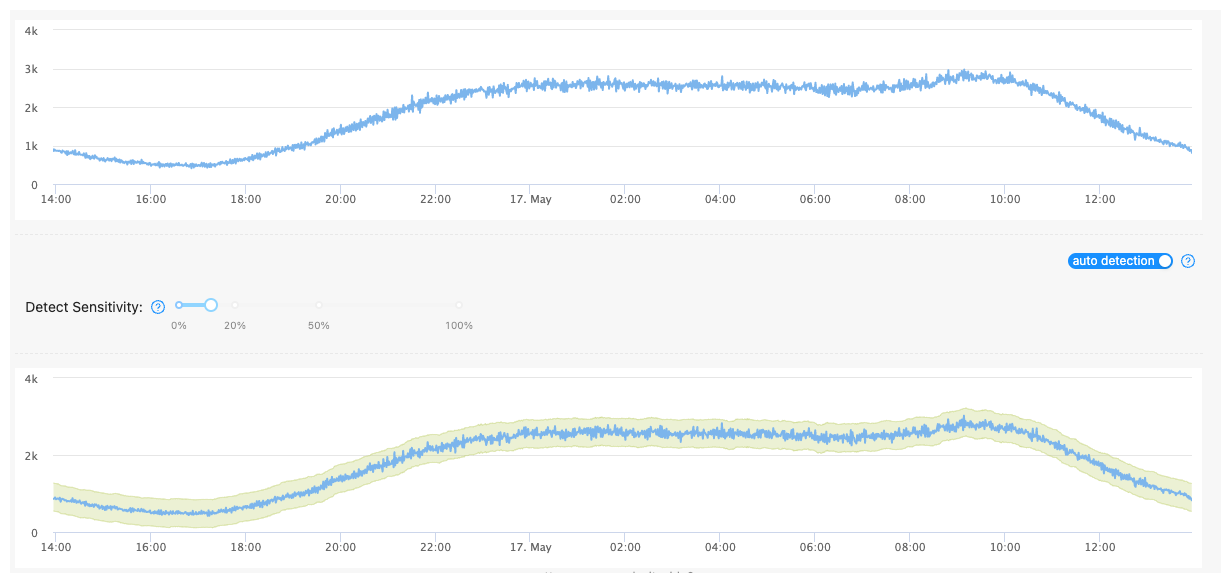}
    \caption{Example of our user interface for automatic time series anomaly detection}
    \label{fig:ui_example}
\end{figure}

To address the aforementioned challenges, we present a comprehensive framework for the automatic optimization of parameters in time series anomaly detection models, irrespective of the model type. Our framework introduces three optimization targets: prediction score, shape score, and sensitivity score.
The prediction score guides the optimization process for prediction-based methods, while the shape score evaluates the visual shape of the detection results. The sensitivity score measures whether the model's detection performance aligns with the user's expected number of anomaly points.
The framework can seamlessly adapt to new model backbones or new time series data. It accomplishes this by optimizing either one or multiple optimization targets, without necessitating prior knowledge or manual labeling efforts.
Through extensive evaluations and real-world deployment for over six months, our framework has demonstrated remarkable results. From the user's perspective, our framework simplifies the process by requiring only a sensitivity value, as illustrated in Figure~\ref{fig:ui_example}, enabling fully automated time series anomaly detection without manual intervention. We also offer a user-friendly fine-tuning interface with a small set of easily understandable parameters. Currently, our automatic time series anomaly detection framework is the most widely utilized algorithm in our monitoring platform, effectively serving over 50,000 time series every minute.
In summary, our work makes the following key contributions:
\begin{itemize}
    \item We formalize the optimization targets for parameter tuning, namely the prediction score, shape score, and sensitivity score. This enables automatic optimization of different model backbones by focusing on one or multiple targets.
    \item We introduce the shape score as a novel optimization target, evaluating the performance of time series anomaly detection based on intuitive observations. Our framework facilitates achieving the most appropriate detection shape for anomaly detection results. We also provide a user-friendly fine-tuning function with a small set of simple parameters that are easily understandable by users. This fine-tuning feedback serves as a valuable resource to improve our shape score model.
    \item We have implemented our proposed framework online, effectively serving multiple user groups and processing over 50,000 time series data every minute. The effectiveness of our approach is evident from the online deployment, where users effortlessly obtain desired detection results by providing a sensitivity value.
\end{itemize}

\section{Related Work}
AutoML has gained significant popularity in machine learning to enhance performance metrics. In the realm of time series analysis, AutoML techniques have been employed to automate various aspects, such as data cleaning and filling missing points~\cite{shende2022cleants} from the data side, as well as model selection and parameter tuning~\cite{chatterjee2022mospat} from the model side. However, in this study, we specifically concentrate on the automation of parameter tuning for time series anomaly detection.

One approach involves directly optimizing the model parameters based on prediction errors. For example, Prophet~\cite{taylor2018forecasting} provides a function to optimize the model's parameters using metrics like root mean squared error (RMSE) and mean absolute percentage error (MAPE). However, optimizing a model solely based on prediction errors can cause it to fit to every point, including anomalies, resulting in an increased number of false negatives. 
Another approach, exemplified by TODS~\cite{lai2021tods}, treats anomaly detection as a binary classification problem and optimizes parameters using anomaly labels, typically relying on metrics such as F1 score or precision. However, this kind of approach is challenging to apply to cases without labeled data, making it impractical for real-world monitoring services where obtaining all the necessary labels beforehand is impossible. 
The last kind of approach trains a model to directly predict the best hyperparameters for a given method. This approach has been successfully employed by industry leaders such as Microsoft~\cite{ying2020automated} and Facebook~\cite{zhang2021self}. For instance, Ying et al.\shortcite{ying2020automated} utilized a LightGBM\cite{ke2017lightgbm} regression model to learn the optimal hyperparameters for various anomaly detection models, enabling the prediction of the best parameter values when encountering new time series. Similarly, Zhang et al.~\shortcite{zhang2021self} employed an offline exhaustive parameter tuning process to determine the best-performing hyperparameters for different model and data combinations. They trained a multi-task neural network, where each task focused on predicting a specific parameter value, which could be either categorical or numerical. It is important to note that this method requires prior knowledge and a significant amount of labeled resources for each algorithm. Furthermore, when introducing a new model, manual parameter tuning is still necessary to generate the correct parameter labels.

To address the challenges mentioned above, we propose three primary optimization targets for automatic univariate time series anomaly detection, which can be seamlessly applied to various model backbones while minimizing the need for extensive labeling efforts.

% TimeAutoAD \cite{jiao2022timeautoad} presents an automatic anomaly detection pipeline to optimize the model configuration and hyperparameters automatically. And use three different strategies to augment the training data for generating pseudo negative time series and employ a self-supervised contrastive loss to distinguish the original time series and the generated time series. 
% Difference: the definition for the anomaly time series in this paper is for the whole time series, and is not for point anomaly. 
% This paper is called automl because of generative training, does not contain the real world needs for automl. Cannot fit to different user's need. 

\section{Optimization Targets}
\begin{figure}[t]
    \centering
    \includegraphics[width=0.85\linewidth]{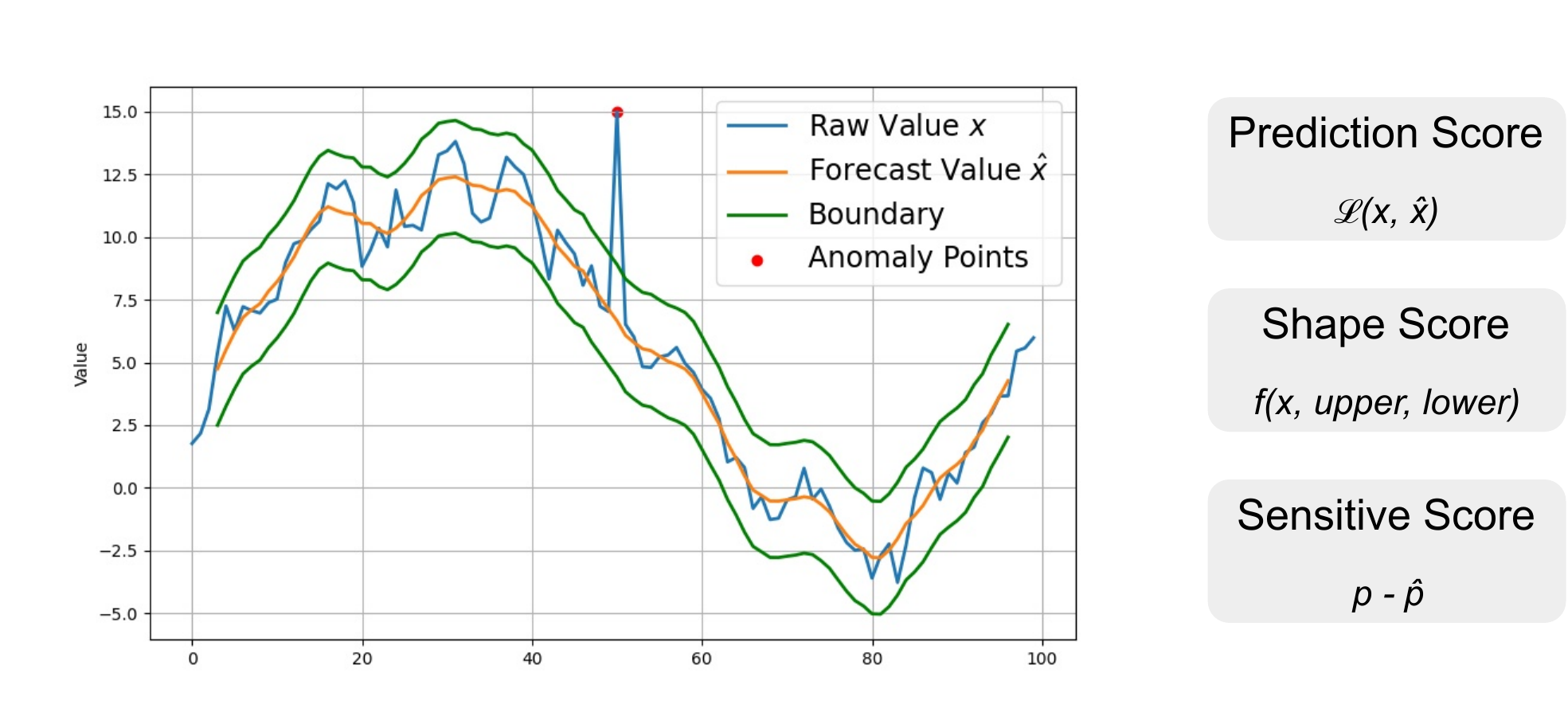}
    \caption{Explanation about the three optimization targets for a forecasting model.}
    \label{fig:optimization_targets}
\end{figure}
We propose three general optimization targets for effectively optimizing the parameters of the anomaly detection model. These optimization targets serve as evaluation metrics to assess the performance of the detection results. 
For example, for forecasting models, the detection output resembles the illustration in Figure~\ref{fig:optimization_targets}. The prediction score evaluates the disparity between the raw time series values $x$ and their corresponding forecasted values $\hat{x}$. On the other hand, the shape score measures the shape of the detection boundary by considering the raw value $x$, the upper boundary $u$, and the lower boundary $l$ as inputs to a shape score model $f$, denoted as $f(x, u, l)$, which outputs a score indicating the performance of the detection boundary. The sensitivity score governs the number of anomalies detected in the results. For instance, if the detected anomalies are represented as $\mathcal{A}$ and a user's desired anomaly proportion is set at $p=1\%$, the model strives to identify a suitable threshold that yields detection results containing approximately 1\% anomaly points, expressed as $\hat{p}=|\mathcal{A}|/|x|$. Therefore, in the case of forecasting models, the sensitivity score directly controls the width of the detection boundary.

It is important to note that not all optimization targets are applicable to every method. For instance, the prediction score is not relevant for methods that do not involve forecasting values. Nonetheless, we demonstrate that nearly all methods can be optimized using at least one of the proposed optimization targets. We will provide further details on setting the optimization targets for different methods. In the subsequent sections, we will outline the specifics for each optimization target.
Overall, we present an automatic parameter tuning framework that tackles the following problem:
\paragraph{Problem definition.}
Given a time series $x$, an anomaly detection model $d$, a set of parameters $\theta$, and a desired sensitivity value $p$, the parameter tuning framework aims to discover the optimal parameter set $\hat{\theta}$ for the model $d$ based on one or multiple optimization targets. These optimization targets consist of the prediction score, shape score, and sensitivity score.

\subsection{Prediction Score}
Prediction score is used to optimize the parameter $\theta$ for a forecasting model $d_\theta$ by minimizing the prediction error: 
\begin{equation}
    min_{\theta} \quad \mathcal{L}(x, d_{\theta}(x))
\end{equation}
where the evaluation metrics are commonly chosen from the followings
\begin{itemize}
\small
    \item Mean absolute error (MAE): $\frac{1}{N}\Sigma^N_{i=1}|x_i - \hat{x}_i|$
    \item Median absolute error (MEDAE): $\text{median}|x_i - \hat{x}_i|$
    \item Root mean squared error (RMSE): $\sqrt{\frac{1}{N}\Sigma^N_{i=1}(x_i - \hat{x}_i)^2}$
    \item Mean absolute percentage error (MAPE): $\frac{100}{N}\Sigma^N_{i=1}|\frac{x_i-\hat{x}_i}{x_i}|$ 
\end{itemize}

In contrast to traditional time series forecasting tasks, prediction-based time series anomaly detection tasks aim to predict the normal pattern rather than every individual point, including anomalies. In real-world time series data, the raw time series may indeed contain anomaly points, such as sudden spikes and dips. However, if a model is trained directly on raw inputs, it may inadvertently learn from both the noise and anomalies present in the data.
Therefore, smoothing strategies are necessary to obtain suitable training prediction targets, allowing the model to learn to fit the normal pattern. Table~\ref{tab:comparison_smoothing} illustrates the performance of the model with different smoothing strategies. The smoothed inputs $\widetilde{x}$ are considered the correct labels for prediction, and the loss is evaluated based on $\mathcal{L}(\widetilde{x}, f_{\theta}(x))$. It is evident that employing simple smoothing strategies, such as filtering extreme values and applying moving averages to each time point, significantly enhances model performance.

\begin{table}[t]
    \centering
    \scriptsize
    \caption{Model performance with different smooth strategies.}
    \label{tab:comparison_smoothing}
    \begin{tabular}{ccccc}
    \hline
        Smooth Strategy & MAE & MEDAE & RMSE & MAPE  \\ \hline
        None   & 59.38$\pm$1.83 & 44.30$\pm$3.34 & 183.56$\pm$8.29 & 89.41$\pm$6.41  \\
        Filter & 48.55$\pm$2.90 & 38.99$\pm$2.21 & 22.95$\pm$2.74 & 64.00$\pm$4.26 \\
        Filter + MA & 44.89$\pm$1.70 & 35.99$\pm$1.31 & 2.93$\pm$0.27 & 59.31$\pm$2.79 \\ \hline
    \end{tabular}
\end{table}

\begin{table}[t]
    \centering
    \scriptsize
    \caption{Model performance on different evaluation metrics with different optimization metrics.}
    \label{tab:comparison_optimization_metric}
    \begin{tabular}{ccccc}
    \hline
        Optimization Metric & MAE & MEDAE & RMSE & MAPE  \\ \hline
        MAE   & 47.73$\pm$1.83 & 37.72$\pm$1.33 & 18.24$\pm$0.24 & 65.59$\pm$2.72  \\
        MEDAE & 47.89$\pm$1.64 & 37.84$\pm$1.24 & 8.09$\pm$0.23 & 64.48$\pm$2.59 \\
        RMSE  & 44.75$\pm$1.63 & 35.57$\pm$1.24 & 5.66$\pm$0.23 & 59.87$\pm$2.53 \\ 
        MAPE  & 44.61$\pm$1.63 & 35.59$\pm$1.25 & 2.11$\pm$0.23 & 59.54$\pm$2.59 \\\hline
    \end{tabular}
\end{table}

Apart from the smoothing strategy, we find that set the correct prediction loss can further enhance the prediction performance. Table~\ref{tab:comparison_optimization_metric} shows the results with using different evaluation metrics as the optimization target. We can see that using MAPE as the optimization target can obtain a model with the best performance. 

In summary, we use two strategies to ensure the optimization process for a prediction model to be resistant to the noisy anomalies, one is use the smoothed inputs as the prediction target, and another is using MAPE as the optimization target for the model training.

\subsection{Shape Score}
\begin{figure}[t]
\centering
\small
\begin{minipage}[t]{0.495\linewidth}
\includegraphics[width=\linewidth]{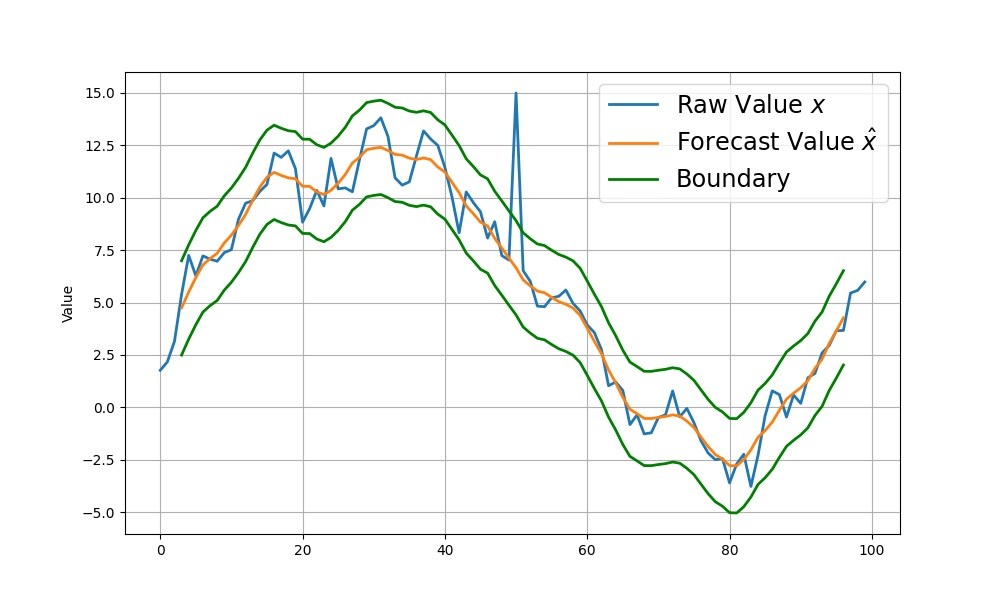}
\centering{(a)}
\end{minipage}
\begin{minipage}[t]{0.495\linewidth}
\includegraphics[width=\linewidth]{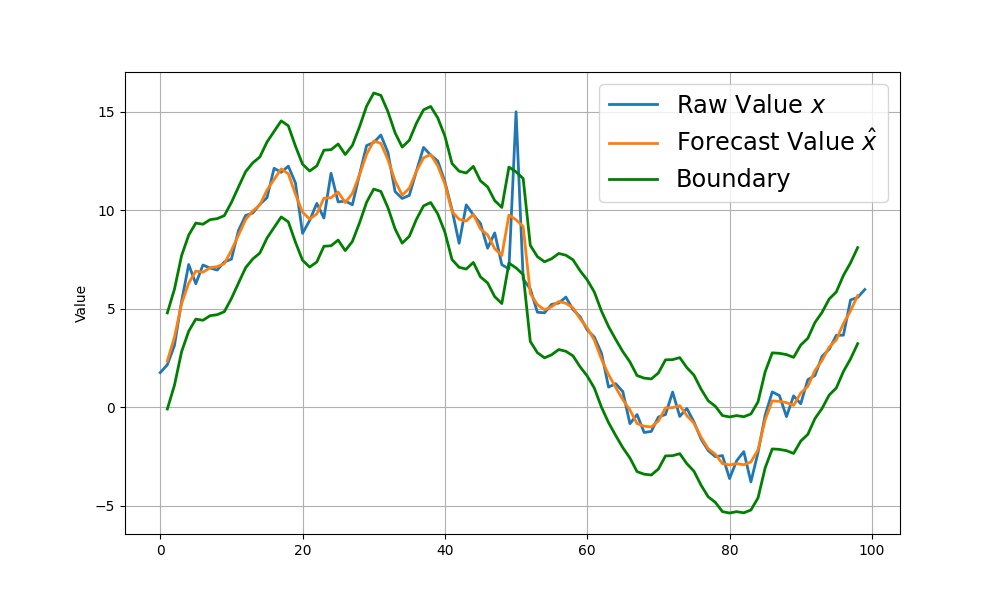}
\centering{(b)}
\end{minipage}
\caption{Example for cases a model has (a) lower prediction score and higher shape score, and (b) higher prediction score and lower shape score.}
\label{fig:example_shape_score}
\end{figure}

The shape score is utilized to evaluate the shape of the detection boundary and determine whether the outputs align with the ideal detection boundary as perceived by humans. 
For instance, in Figure~\ref{fig:example_shape_score}, the left figure illustrates an example where a model has a lower prediction score but a higher shape score. In comparison, the right figure has a higher prediction score but the results appear to be overly fitted to each point, including the anomaly points, indicating a lower shape score. 
Intuitively, we prefer the model shown in the left figure because the spike in the figure is more likely to be an anomaly. Thus, we can say that the shape of the detection results in the left figure is better than the shape of the detection results in the right figure. 
We quantify this intuitiveness as the \textbf{\textit{shape score}} and train a shape score model to capture it.
We formalize the shape score model $f$ to take the raw values $x$ and the boundaries $(u,l)$ as the inputs and produces a single shape score value $\hat{s} = f(x, u, l)$ ranging from 0 to 1, where a higher score indicates better performance in capturing the desired shape of the detection results.
The shape score is used to optimize the parameter $\theta$ for an anomaly detection model $d_\theta$ by maximizing the shape score:
\begin{equation}
    max_\theta \quad f(x, u, l) \quad u=g(d_\theta(x)), l = h(d_\theta(x))
\end{equation}
where $g$ and $h$ are the transformation of the prediction values $d_\theta(x)$.

\paragraph{Dataset Formulation.} To train an effective shape score model, it is crucial to have a high-quality labeled dataset comprising both good and bad detection cases. 
We formulate the base training dataset using the following strategies. 
For the \textbf{good cases}, we employ a combination of data synthesis and manual labeling. Initially, we manually annotate a small subset from our monitoring services and assign shape scores ranging from 0 to 1 based on intuitive observations. For instance, in Figure~\ref{fig:example_shape_score} (a), a shape score of 1 is assigned, while in Figure~\ref{fig:example_shape_score} (b), a shape score of 0.6 is assigned. Additional instances will be marked as good cases via our user fine-tuning service mentioned in Section~\ref{sec:fine_tune_service}. A manual filtering strategy will be used to selectively add examples that exhibit new patterns compared to the existing labeled cases.
In the synthesized dataset, we generate various base patterns ${x_b}$ such as seasonal sine-like curves, sparse inputs, and random-walk-like patterns. To simulate anomalies in real-world monitoring services, we introduce noises and anomalies ${x_a}$ to the base patterns, resulting in the synthesized data $x = x_b + x_a$.
Considering previous observations, we assert that the detection boundary should only learn from the base patterns. Thus, we set the ideal upper boundary as $u = x_b + 3\sigma$ and the ideal lower boundary as $l = x_b - 3\sigma$, where $\sigma$ represents the standard deviation of the inputs. We label the shape score for this set of detection results $(x, u, l)$ as 1, indicating good performance.
For the \textbf{bad cases}, we introduce eight types of anomalies into the detection boundaries of the good cases. These anomalies include inverting the upper and lower boundaries, positioning the lower boundary above the raw values, positioning the upper boundary below the raw values, excessively narrow or broad boundaries, boundaries with numerous high-deviation noises, boundaries with extreme value peaks, and boundaries aligned with significantly changed raw values. The shape scores assigned to these cases are lower than the original shape score, indicating poor performance in capturing the desired shape of the detection results.

\paragraph{Model Structure}
The shape score model takes the raw value $x$ and the boundaries $(u, l)$ as inputs. In real-world scenarios, the length $N$ of the time series $x$ can vary from days to weeks, requiring the shape score model to handle time series of different sizes.
One approach is to train a shape score model that can handle inputs with a fixed window size $W$. If a time series is longer than this window, it is divided into multiple windows to obtain individual shape scores, and the sum of these scores is used as the final shape score. However, this approach has a limitation in that the model can only focus on the shape score within a specific window, potentially missing anomalies that are only detectable when considering the entire time series.
To address this issue, we need to reduce or increase the dimensionality of the time series to ensure consistent input dimensions for the shape score model. A suitable approach is to transform the input into images since the shape score is also based on visual observations. Specifically, we utilize GASF (Gramian Angular Summation Field) \cite{wang2015imaging} to represent the time series as an image.
The core idea of GASF is to first use Piecewise Aggregation Approximation (PAA) \cite{keogh2000scaling} to smooth the time series while preserving its trends and reducing its size. Next, the reduced time series is projected onto a polar coordinate system, ensuring a bijective transformation that preserves the information. As a result, the inputs $(x, u, l)$ can be transformed into three 2-dimensional images, forming a composite image with three layers. This allows us to utilize deep learning frameworks, with CNN (Convolutional Neural Network) chosen as the model backbone, to learn the shape score.

% \begin{figure}[t]
%     \centering
%     \includegraphics[width=\linewidth]{figures/example_shape_score_model.pdf}
%     \caption{Structure of the shape score model}
%     \label{fig:shape_score_model}
% \end{figure}

The optimization process for the shape score model involves minimizing the loss function $\mathcal{L}(s, f(x, u, l))$. It is important to note that while the shape score model is still learned through supervised learning, our proposed method offers a distinct advantage compared to the methods described in \cite{lai2021tods} and \cite{zhang2021self}.
Our approach presents a general model that can be applied to various methods and different time series datasets without the need for additional labeling efforts.

\subsection{Sensitive Score}
Real-world monitoring services typically involve the monitoring of millions of time series data. In such scenarios, it becomes challenging for users to manually label all anomalies present in the data. Moreover, users may have varying degrees of sensitivity towards anomalies. Some users may prioritize capturing every possible anomaly, while others may focus only on the most extreme anomalies.

To accommodate these preferences, we introduce the notion of an anomaly ratio denoted as $p$. The anomaly ratio represents the user's desired proportion of anomalies in the detection results. For instance, if a user sets $p=0.05$, it indicates that they expect the detection results to contain approximately 5\% of the total anomalies present in the data. By adjusting the anomaly ratio, users can customize the sensitivity of the model to align with their specific requirements and priorities.
\begin{figure}[t]
\centering
\begin{minipage}[t]{0.47\linewidth}
\includegraphics[width=\linewidth]{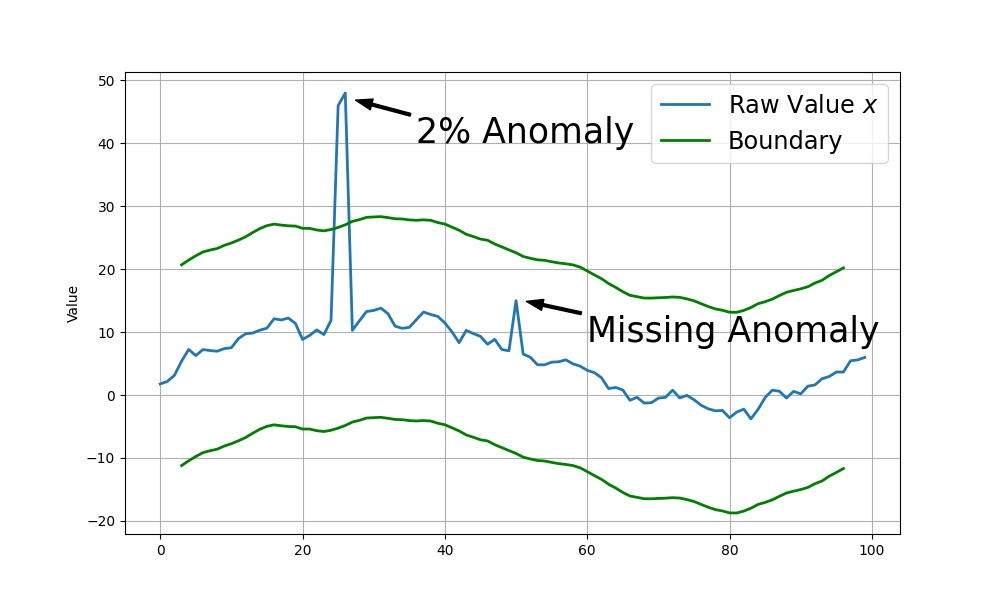}
\centering{(a) Missing anomalies}
\end{minipage}
\begin{minipage}[t]{0.47\linewidth}
\includegraphics[width=\linewidth]{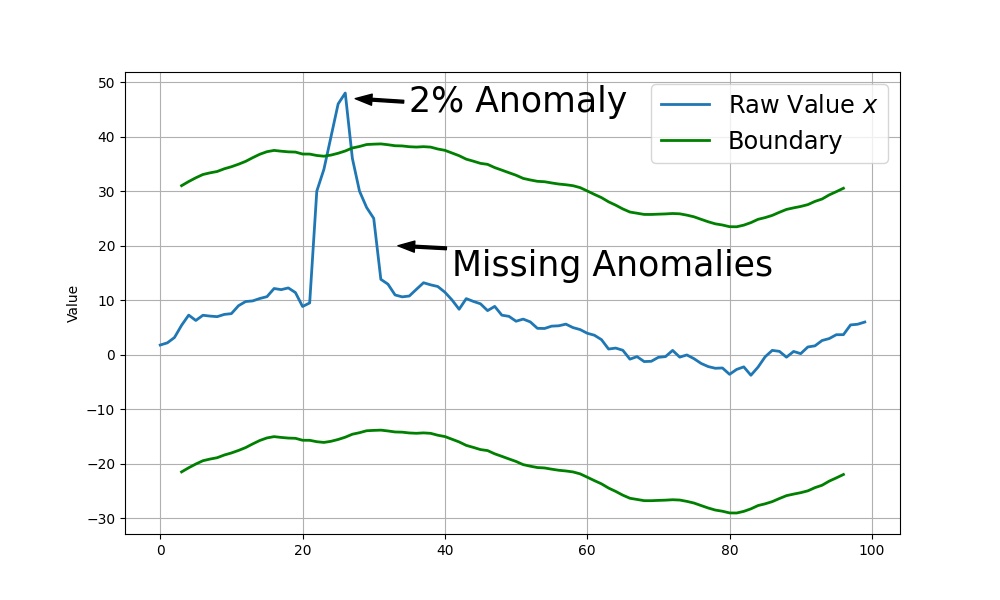}
\centering{(b) Too broad boundary}
\end{minipage}
\caption{Examples for the tuned threshold based on the given anomaly ratio only.}
\label{fig:example_issue_rate}
\end{figure}

% \vspace{-0.5cm}

\begin{figure}[t]
\centering
\begin{minipage}[t]{0.47\linewidth}
\includegraphics[width=\linewidth]{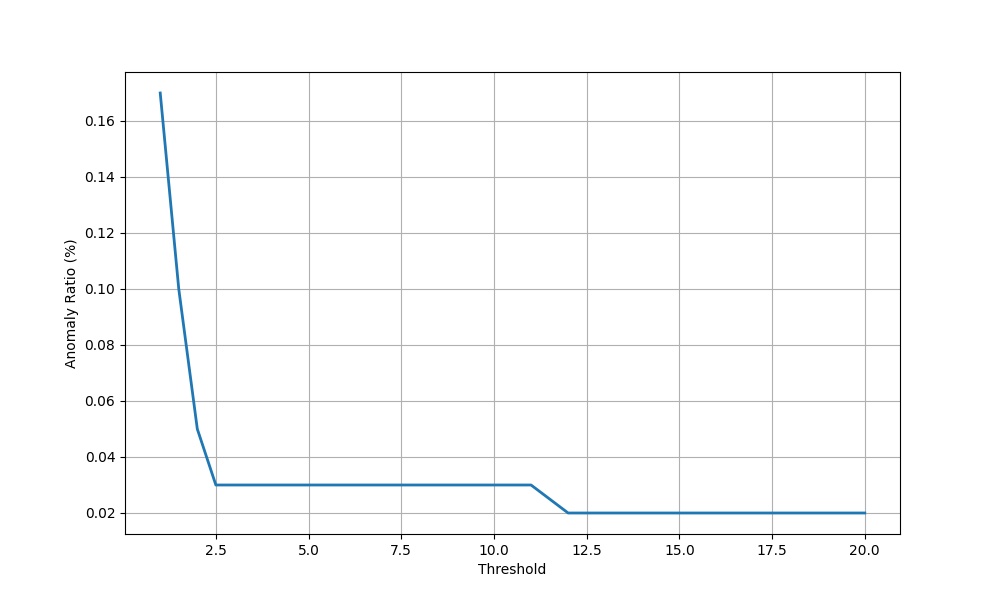}
\centering{(a)}
\end{minipage}
\begin{minipage}[t]{0.47\linewidth}
\includegraphics[width=\linewidth]{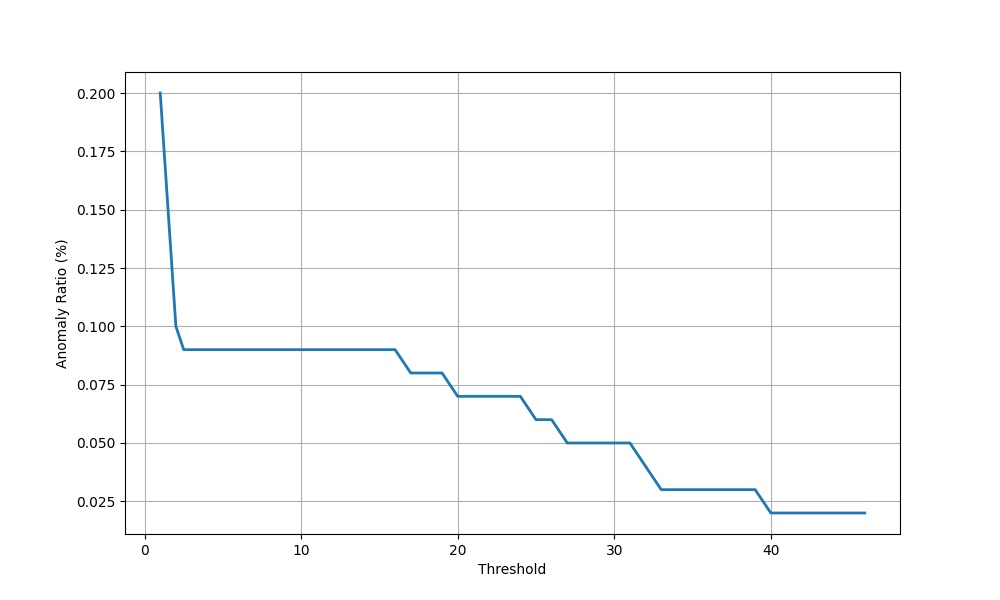}
\centering{(b)}
\end{minipage}
\caption{The relationship between threshold value and anomaly ratio given the time series shown in Figure~\ref{fig:example_issue_rate}.}
\label{fig:example_tuning}
\end{figure}
The most straightforward approach to tune a model based on the anomaly ratio $p$ is to search for a threshold that precisely satisfies the desired ratio. However, this approach can result in two potential cases that lead to false negatives, as illustrated in Figure~\ref{fig:example_issue_rate}.
In Figure~\ref{fig:example_issue_rate}(a), the missed peak point within the detection boundary could also be considered an anomaly. Similarly, in Figure~\ref{fig:example_issue_rate}~(b), the points near the top 2\% of anomaly points should also be identified as anomalies.
To address this issue, it is crucial to consider the relationship between the threshold value and its corresponding anomaly ratio. Figure~\ref{fig:example_tuning} illustrates this relationship given the time series data from Figure~\ref{fig:example_issue_rate}.
We observe that increasing the threshold value at certain intervals can significantly reduce the detected anomaly ratio, while in other cases, increasing the threshold has only a slight effect on reducing the ratio.
Based on this observation, we can determine the threshold value by identifying either: 1) the starting point with lower derivatives or 2) the active points that can significantly reduce the sensitivity values. This transforms our problem into finding the threshold values for the knee points and active points in a given time series.
To achieve this, we employ the methods proposed in \cite{satopaa2011finding} to detect the knee points, and we identify the active points by selecting those that contribute the most to the decrease in the anomaly ratio.
As a result, we obtain a set of threshold values denoted as $\{t|t\in T\}$, and the corresponding anomaly ratio for a given detection result is $p_t = P(f(x)|t)$. Our objective is to find a threshold that minimizes the distance to the user-defined anomaly ratio $p$. We refer to this objective as the sensitivity score:
\begin{equation}
\min_{t} |p - p_t| \quad \text{s.t.} \quad \{t|t\in T\}
\end{equation}

% \cite{laptev2015generic}: yahoo automl. Does not contain information for model selection and parameter tuning. As for the threshold selection, it is based on the anomaly score. The anomalies are determined by statistical significance or by some clustering based methods. 
% Boundary tuning. \cite{umsonst2022finite}
% Tuned based on order statistics to estimate the detector threshold. 
% They formulate the problem as: they want to find the smallest threshold $J_{D}$ such that prob$\{y_D <= J_D\} >= \gamma$, the threshold $J_D$ will result in a false alarm probability of at most $1-\gamma$. 
% They propose solutions based on three types of statistical methods: the DKW inequality, the Chebyshev's inequality, and the finite guarantees from exact confidence intervals of the binomial distribution. 

\section{The Automation Framework at eBay}
\begin{figure}[t]
    \centering
    \includegraphics[width=0.8\linewidth]{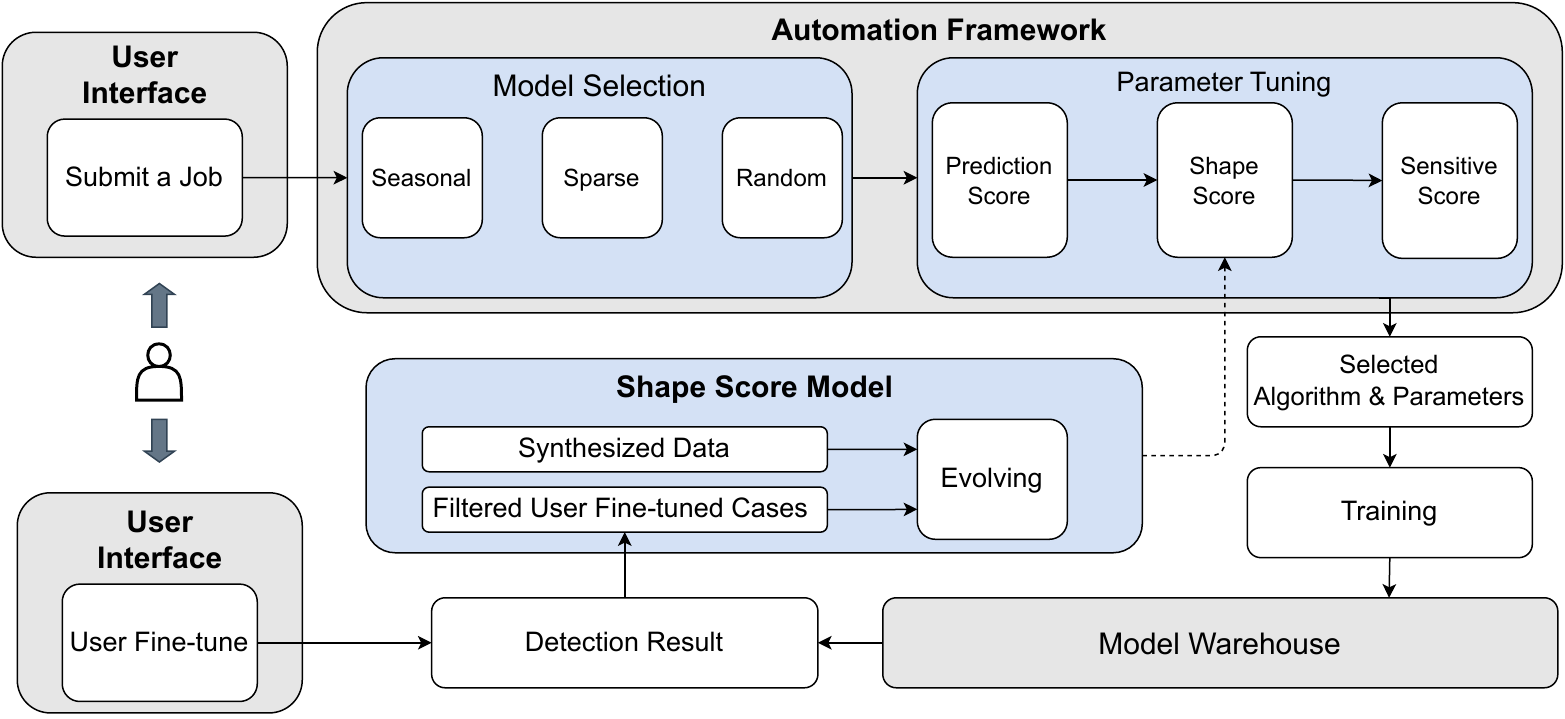}
    \caption{The automation framework at eBay}
    \label{fig:automation_framework}
\end{figure}

At eBay, we have successfully integrated our proposed parameter tuning targets into our existing platform to optimize detection algorithm parameters (Figure~\ref{fig:automation_framework}).
When a user submits a detection job, the data dumper retrieves the relevant time series data. Our automation framework uses a trained LightGBM~\cite{ke2017lightgbm} classifier to identify patterns, such as seasonality, sparsity, or randomness, and selects the appropriate detection model.
The parameter tuning framework then determines optimal parameters for the selected method, categorized into prediction score, shape score, and sensitive score. The optimization is conducted sequentially, targeting each objective. If a method lacks parameters for a specific target, we optimize using the other targets. Overlapping parameters between prediction score and shape score are optimized based on their combined score.
Once the algorithm and tuned parameters are determined, they are applied to the detection flow, generating the final results.
In cases where users require further fine-tuning, their feedback becomes a valuable resource for updating the shape score model.

\subsection{Parameter Tuning for Different Methods}
As mentioned earlier, we consider three types of time series patterns: seasonal, sparse, and random patterns. We employ three corresponding algorithms to detect anomalies in each pattern.
\begin{itemize}
    \item \textbf{Random Pattern}. For random-walk-like patterns, we utilize a moving average method, similar to the approach proposed in \cite{yu2016improved}, to detect anomalies. The prediction for the next point is based on the \texttt{average} value (mean or median) within a specified time window with size \texttt{window\_size}, and the anomaly boundary is determined using a \texttt{threshold}-sigma calculation on the variation. To optimize this method, we first select the parameters for \texttt{average} and \texttt{window\_size} based on the shape score. Then, we optimize the \texttt{threshold} using the sensitive score.
    \item \textbf{Sparse Pattern}. In the case of sparse time series, we focus solely on extreme anomaly values. To detect such anomalies, we employ Extreme Value Theory (EVT) \cite{siffer2017anomaly}. This method produces upper and lower boundaries without forecasting values. Two parameters control the initial location of the boundary: the first parameter truncates the original data distribution to focus on the higher values, while the second parameter sets the initial expected anomaly ratio. To optimize this method, we first tune these two parameters based on the shape score. Subsequently, we optimize another parameter, the threshold that directly controls the boundary, using the sensitive score. 
    \item \textbf{Seasonal Pattern}. For seasonal time series data, we utilize a seasonal decomposition method \cite{li2020anomaly} to detect anomalies. The prediction for a point is based on its value within the same past seasonal window, as well as values from the past time window. Parameters controlling the window sizes of the seasonal, trend, and residual windows are tuned based on the prediction score. The anomaly boundary for this method is determined by the distribution of prediction residuals. Additionally, there is a threshold parameter that controls the width of the boundary, which we optimize using the sensitive score.
\end{itemize}

\subsection{Evaluation on eBay's monitoring service}
We evaluate the parameter tuning performance using eBay's monitoring dataset. The dataset comprises 50 time series collected from eBay's production environment over the past month, with each time series representing minute-level data.
In our evaluation, we employed the model selection algorithm that determines the most suitable algorithm for each time series based on their unique pattern features. Subsequently, we assessed the performance before and after applying the tuned parameters.
To evaluate the effectiveness of the tuning process, we utilized two widely-used evaluation metrics: the point-wise F1 score and the AUC measure, as mentioned in \cite{paparrizos2022tsb}. 
\begin{table}[h]
    \centering
    \small
    \caption{Evaluation results: before and after parameter tuning}
    \begin{tabular}{ccccc}
    \toprule
    \multirow{2}{3em}{Method}     & \multicolumn{2}{c}{F1} & \multicolumn{2}{c}{AUC} \\ 
                    & Before    & After   & Before   & After \\ \midrule
    % Random method   & 0.404 & 0.716 & 0.874 & 0.965 \\
    % Sparse method   & 0.458 & 0.929 & 0.905 & 0.957    \\
    % Seasonal method & 0.507 & 0.692 & 0.902 & 0.965    \\
    % Overall         & 0.446 & 0.730 & 0.883 & 0.964    \\
    Random method   & 0.404 & 0.874\textcolor{teal}{$\uparrow$} & 0.716 & 0.965\textcolor{teal}{$\uparrow$} \\
    Sparse method   & 0.458 & 0.905\textcolor{teal}{$\uparrow$} & 0.929 & 0.957\textcolor{teal}{$\uparrow$}    \\
    Seasonal method & 0.507 & 0.902\textcolor{teal}{$\uparrow$} & 0.692 & 0.965\textcolor{teal}{$\uparrow$}    \\
    % Overall         & 0.446 & 0.730 & 0.883 & 0.964    \\
    \bottomrule
    \end{tabular}
    \label{tab:results_on_ebay}
\end{table}
Table~\ref{tab:results_on_ebay} presents the evaluation results for the time series classified into the three methods, as well as the overall performance. The findings clearly demonstrate that our proposed parameter tuning methods have a significant positive impact on the performance of the algorithms.

\subsection{User Fine-tuning Service}
\label{sec:fine_tune_service}
\begin{figure}[t]
    \centering
    \includegraphics[width=0.9\linewidth]{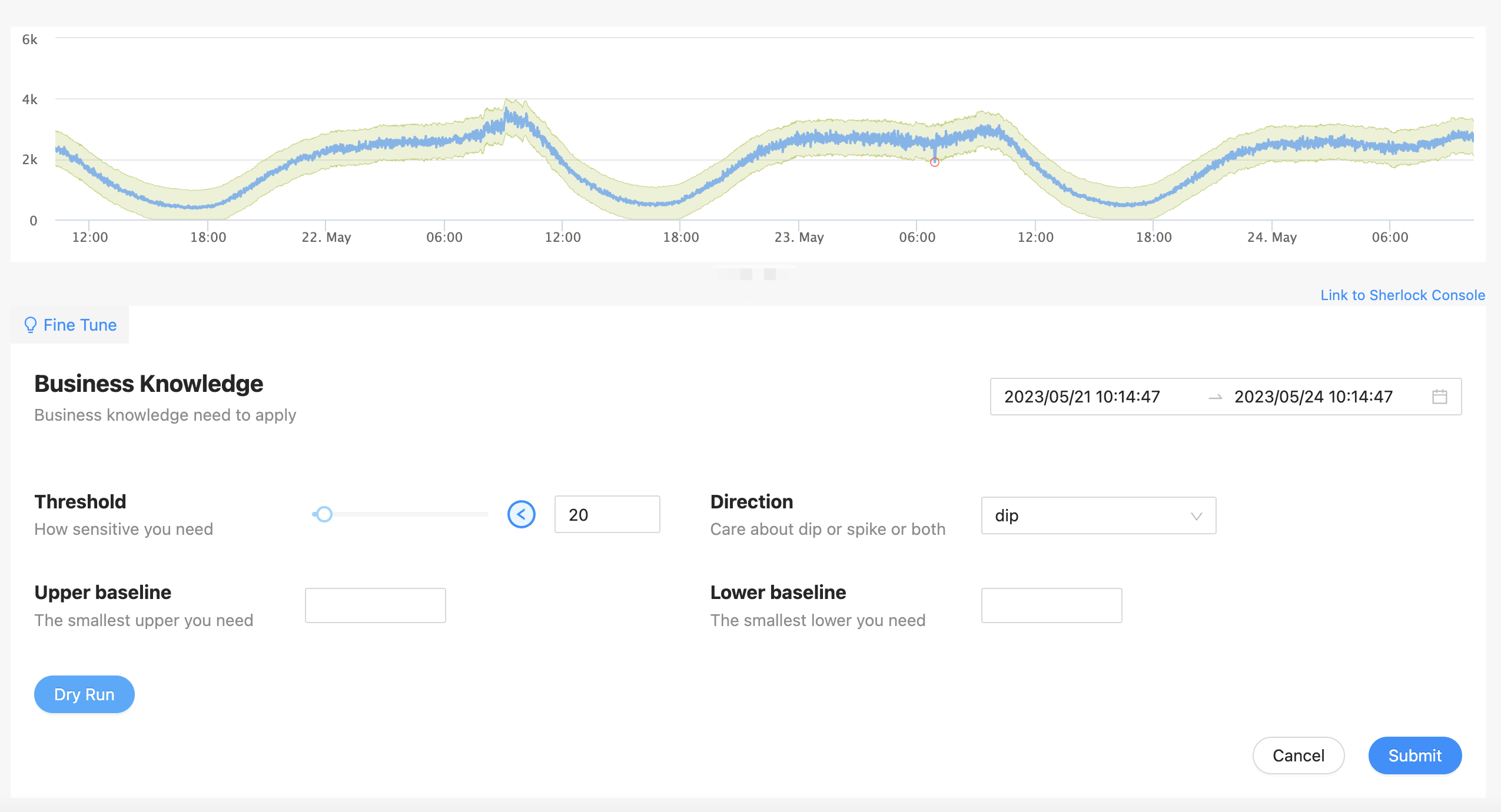}
    \caption{Our user interface for the fine-tuning service.}
    \label{fig:fine_tuning_service}
\end{figure}

In some cases, customers may still find the detection results unsatisfactory even after tuning the parameters. This can happen due to two reasons: firstly, the trained model may lack the necessary business knowledge, and secondly, there may be new cases that are not covered by the model. To address these situations, we offer a user-friendly fine-tuning service that enables users to directly adjust the detection results, which serves as valuable training cases for the shape score model. Specifically, examples that exhibit new patterns compared to the existing labeled cases are added to the training dataset. This enhances the model's ability to handle novel cases.
Our fine-tuning service exposes four parameters, illustrated in Figure~\ref{fig:fine_tuning_service}. The first parameter is the threshold, which controls the width of the detection boundary. The second parameter is the upper baseline, where values below this line are not considered anomalies. Similarly, the third parameter, the lower baseline, ensures that values above this line are not classified as anomalies. The fourth parameter, called direction, allows users to specify the side of the anomalies they want to focus on.
The fine-tuning process is based on the loaded model cache and does not require any additional training, making it a quick operation that can be completed within seconds.
By providing these user-defined parameters and obtaining the updated detection results, we can further refine the shape score model through additional training. This iterative process enables continuous improvement and adaptation to specific user requirements and evolving anomaly patterns.

\section{Extensional Experiments}
To further evaluate the performance of the proposed parameter tuning framework, we conduct experiments on several public datasets that also used to monitoring services and test the parameter tuning performance on other time series anomaly detection methods. 

% \paragraph{Datasets.} We evaluate the parameter tuning performance on the following public datasets:
% \textbf{IOPS}~\footnote{\url{http://iops.ai/}}: This dataset includes multiple indicators reflecting the scale, quality of web services, and machine health status. It consists of 58 time series, each with an average length of approximately 100,000 data points. The anomalies in each time series account for around 2\% of the data.
% \textbf{NAB}~\cite{ahmad2017unsupervised}: NAB dataset comprises labeled real-world and artificial time series, such as server metrics and traffic data. It contains 58 time series with an average length of 6,000 data points, and anomalies make up about 9\% of the data.
% \textbf{Yahoo}~\footnote{\href{https://webscope.sandbox.yahoo.com/catalog.php?datatype=s&did=70&guccounter=1}{Yahoo dataset.}}: Published by Yahoo Labs, this dataset consists of real and synthetic time series based on production traffic data from Yahoo's systems. It includes 367 time series with an average of 1,561 data points, and the anomaly ratio is approximately 0.7\%.
% \textbf{SMD}~\cite{su2019robust}: SMD dataset is a 5-week-long dataset collected from a large Internet company. It comprises three groups of entities from 28 different machines, resulting in a total of 281 time series. Each time series contains an average of 25,562 data points, and the anomaly ratio is around 3.52\%.

\paragraph{Datasets.} We evaluate parameter tuning on the following public datasets:
\textbf{IOPS}~\footnote{\url{http://iops.ai/}}: 58 time series reflecting web service indicators, machine health, and scale. Average length: 100,000 data points. Anomaly ratio: 2\%.
\textbf{Yahoo}~\footnote{\href{https://webscope.sandbox.yahoo.com/catalog.php?datatype=s&did=70&guccounter=1}{Yahoo dataset.}}: 367 real and synthetic time series based on production traffic. Average length: 1,561 data points. Anomaly ratio: 0.7\%.
\textbf{SMD}~\cite{su2019robust}: 5-week dataset from a large Internet company. 281 time series from 28 machines. Average length: 25,562 data points. Anomaly ratio: 3.52\%.

% \begin{table}
%     \centering
%     \footnotesize
%     \caption{Statistics of the datasets}
%     \begin{tabular}{cccc}
%     \toprule
%         Dataset & \# Time Series & Avg Length & $\%$ Anomalies   \\ \midrule
%         IOPS & 58 & 102119 & 2.26 \\
%         NAB & 58 & 6301 & 9.13 \\ 
%         SMAP & 54 & 8066 & 12.80 \\
%         MSL & 27 & 2731 & 10.48 \\
%         Yahoo & 367 & 1561 & 0.69 \\
%         SMD & 281 & 25562 & 3.52 \\ 
%         \bottomrule
%     \end{tabular}
%     \label{tab:dataset_stats}
% \end{table}

\begin{table*}[t]
    \centering
    \scriptsize
    \caption{Experiments before/after using our proposed optimization targets.}
    \begin{tabular}{ccclllllllllll}
    \toprule
        Dataset & Metrics & Method & MA    & EVT   & Elastic & Prophet & LOF   & DBSCAN & CNN   & LSTM  &   AE  & D-Linear & Informer \\ 
        \midrule
        IOPS    & F1      & Before & 0.171 & 0.266 & 0.282   & 0.311   & 0.080 & 0.140  & 0.273 & 0.328 & 0.130 & 0.241    & 0.307 \\
                &         & After  & 0.195\textcolor{teal}{$\uparrow$} & 0.270\textcolor{teal}{$\uparrow$} & 0.287\textcolor{teal}{$\uparrow$} & 0.312\textcolor{teal}{$\uparrow$} & 0.095\textcolor{teal}{$\uparrow$} & 0.150\textcolor{teal}{$\uparrow$} & 0.282\textcolor{teal}{$\uparrow$} & 0.357\textcolor{teal}{$\uparrow$} & 0.309\textcolor{teal}{$\uparrow$} & 0.277\textcolor{teal}{$\uparrow$} & 0.311\textcolor{teal}{$\uparrow$} \\
                & AUC     & Before & 0.659 & 0.648 & 0.789 & 0.772     & 0.500 & 0.580  & 0.774 & 0.795 & 0.630 & 0.745 & 0.782 \\
                          && After  & 0.696\textcolor{teal}{$\uparrow$} & 0.672\textcolor{teal}{$\uparrow$} & 0.794\textcolor{teal}{$\uparrow$} & 0.778\textcolor{teal}{$\uparrow$} & 0.644\textcolor{teal}{$\uparrow$} & 0.700\textcolor{teal}{$\uparrow$} & 0.788\textcolor{teal}{$\uparrow$} & 0.811\textcolor{teal}{$\uparrow$} & 0.804\textcolor{teal}{$\uparrow$} & 0.782\textcolor{teal}{$\uparrow$} & 0.806\textcolor{teal}{$\uparrow$} \\ \midrule
        Yahoo   & F1      & Before & 0.240 & 0.063 & 0.598 & 0.573     & 0.110 & 0.050  & 0.454 & 0.477 & 0.060 & 0.607 & 0.534 \\
                          && After  & 0.243\textcolor{teal}{$\uparrow$} & 0.093\textcolor{teal}{$\uparrow$} & 0.646\textcolor{teal}{$\uparrow$} & 0.575\textcolor{teal}{$\uparrow$} & 0.085\textcolor{purple}{$\downarrow$} & 0.040\textcolor{purple}{$\downarrow$} & 0.479\textcolor{teal}{$\uparrow$} & 0.491\textcolor{teal}{$\uparrow$} & 0.065\textcolor{teal}{$\uparrow$} & 0.666\textcolor{teal}{$\uparrow$} & 0.561\textcolor{teal}{$\uparrow$} \\
                & AUC     & Before & 0.928 & 0.696 & 0.964 & 0.902     & 0.860 & 0.670  & 0.822 & 0.873 & 0.790 & 0.930 & 0.915 \\
                          && After  & 0.930\textcolor{teal}{$\uparrow$} & 0.700\textcolor{teal}{$\uparrow$} & 0.974\textcolor{teal}{$\uparrow$} & 0.904\textcolor{teal}{$\uparrow$} & 0.862\textcolor{teal}{$\uparrow$} & 0.630\textcolor{purple}{$\downarrow$} & 0.839\textcolor{teal}{$\uparrow$} & 0.886\textcolor{teal}{$\uparrow$} & 0.809\textcolor{teal}{$\uparrow$} & 0.941\textcolor{teal}{$\uparrow$} & 0.921\textcolor{teal}{$\uparrow$} \\ \midrule
        SMD     & F1      & Before & 0.178 & 0.162 & 0.240 & 0.235     & 0.180 & 0.360  & 0.267 & 0.257 & 0.090 & 0.283 & 0.266 \\
                          && After  & 0.185\textcolor{teal}{$\uparrow$} & 0.172\textcolor{teal}{$\uparrow$} & 0.245\textcolor{teal}{$\uparrow$} & 0.247\textcolor{teal}{$\uparrow$} & 0.184\textcolor{teal}{$\uparrow$} & 0.360- & 0.269\textcolor{teal}{$\uparrow$} & 0.255\textcolor{purple}{$\downarrow$} & 0.152\textcolor{teal}{$\uparrow$} & 0.296\textcolor{teal}{$\uparrow$} & 0.265\textcolor{purple}{$\downarrow$} \\
                & AUC     & Before & 0.570 & 0.625 & 0.676 & 0.757     & 0.690 & 0.700  & 0.751 & 0.753 & 0.630 & 0.762 & 0.751 \\
                          && After  & 0.597\textcolor{teal}{$\uparrow$} & 0.624\textcolor{purple}{$\downarrow$} & 0.677\textcolor{teal}{$\uparrow$} & 0.754\textcolor{purple}{$\downarrow$} & 0.695\textcolor{teal}{$\uparrow$} & 0.710\textcolor{teal}{$\uparrow$} & 0.754\textcolor{teal}{$\uparrow$} & 0.749\textcolor{purple}{$\downarrow$} & 0.658\textcolor{teal}{$\uparrow$} & 0.772\textcolor{teal}{$\uparrow$} & 0.752\textcolor{teal}{$\uparrow$} \\
        \bottomrule
    \end{tabular}
    \label{tab:overall_performance}
\end{table*}

\paragraph{Methods.} 
In addition to the previously mentioned moving average (MA) method and extreme value theory (EVT) method, we evaluate several other public methods for time series anomaly detection. We examine these methods both with and without utilizing our proposed parameter tuning optimization targets. However, it is important to note that we do not include the results for Matrix Profile~\cite{yeh2016matrix} due to its limited parameters, such as the time window, which can be easily determined through frequency analysis.
\begin{itemize}
% \small
    \item Elastic Net~\cite{zou2005regularization} is a prediction-based anomaly detection algorithm that combines linear regression with L1 and L2 loss. The ratio of the two losses and the input window size are crucial parameters. We use the shape score to determine the optimal parameter combinations.
    \item Prophet~\cite{taylor2018forecasting} is a decomposition-based time series forecasting algorithm. Two parameters, namely the expected change points ratios and the input window size, influence the shape of the detection results. We tune these parameters based on the shape score.
    \item Local Outlier Factor (LOF)~\cite{breunig2000lof} is a clustering-based anomaly detection approach that assigns a binary label to each point based on values within a specified window. The window size and the number of neighborhoods are two parameters with a significant impact on the detection results. We optimize these parameters step by step using the sensitive score.
    \item DBSCAN~\cite{ester1996density} is a clustering-based anomaly detection method. The sliding window length (controlling the number of values for calculation), epsilon (the radius of a circle), and min points (the minimum number of points in a circle) are key factors that affect the detection results. We tune each parameter individually using the sensitive score. 
    \item CNN~\cite{munir2018deepant} is a forecasting-based approach for anomaly detection, where the shape score is used to determine the appropriate kernel size and stride.
    \item LSTM~\cite{hochreiter1997long} is a forecasting-based approach for anomaly detection. The size of the hidden units and the number of neural layers are two crucial hyperparameters that impact the prediction performance. We utilize the shape score to search for the optimal combinations.
    \item Autoencoder (AE)~\cite{sabokrou2015real} utilizes the reconstruction error to detect anomalies. The size of the sliding window is a significant hyperparameter that affects the shape of the reconstructed inputs. Therefore, we employ the shape score to find the best value for this hyperparameter.
    \item D-Linear~\cite{zeng2022transformers} is a simple one-layer neural prediction model. The input sequence size and the forecasting window size are two key factors that impact the performance. We use the shape score to select a suitable value for these parameters.
    \item Informer~\cite{zhou2021informer} is a transformer-based time series forecasting model that encodes the inputs to hidden units and directly predicts the output by feeding masked inputs. The input sequence size, the number of encoder and decoder layers, and the dimension of the hidden units are key hyperparameters that contribute to the final results. We tune them using the shape score. 
\end{itemize}

\paragraph{Results}
We split the data into training and testing sets, using the initial 30\% of the data for training and the remaining for testing. For methods that use the sensitive score, we set the expected anomaly ratio in the training data to match the actual anomaly ratio. If there are no anomaly points in the training data, we use a default ratio of 1\%.
In general, we observe that using the proposed parameter tuning optimization targets improves the detection performance. However, the extent of improvement varies depending on the dataset and the method used. There are two main reasons for these variations. Firstly, when the methods are applied individually on the shape score trained by our production datasets, the public datasets may exhibit different patterns. Secondly, some methods can only learn based on a single sensitive score, thereby missing out on the benefits from the other optimization targets.

% The experiment concludes DBSCAN also benefits from the proposed framework while the observed advantages are inherently constrained as neither prediction score nor shape score is applicable. 

% Comprehensively, the dataset is prepared by the Min-Max scale in the range between 0 and 1. If not specified the dataset itself, the initial 30\% of the data is allocated for training and the remaining is for testing. The optimization framework proposed is used to identify the appropriate hyper-parameter group by conducting 10 trials on the training data.
% The issue rate is generated from the training data label. If the training data comprises of the normal only, 0.01 is used.

\section{Conclusion}
In conclusion, our proposed comprehensive framework for automatic parameter optimization in time series anomaly detection on monitoring services offers three optimization targets: the prediction score, the shape score, and the sensitivity score. 
Through extensive evaluations and real-world deployment, our framework has showcased remarkable results, effectively reducing the need for manual expert fine-tuning and streamlining the detection process for users. 
% By leveraging the optimization targets, we have achieved significant improvements in detection performance, enhancing accuracy and reliability in identifying anomalies. 
However, it is important to acknowledge that the effectiveness of the framework may vary depending on the dataset and algorithm employed. 
Factors such as dataset characteristics and algorithmic limitations can impact the performance of the parameter tuning methods.
Further research is warranted to explore additional optimization targets and their applicability to enhance time series anomaly detection in diverse scenarios.

% \color{gray}

% \section{Style and Format}

% \subsection{Title and Author Information}

% \subsubsection{Author Names}

% Each author name must be followed by:
% \begin{itemize}
%     \item A newline {\tt \textbackslash{}\textbackslash{}} command for the last author.
%     \item An {\tt \textbackslash{}And} command for the second to last author.
%     \item An {\tt \textbackslash{}and} command for the other authors.
% \end{itemize}

% \subsection{Special Sections}
% \subsubsection{Contribution Statement}

% Use
% \begin{quote}
%     {\tt \textbackslash{}section*\{Contribution Statement\}}
% \end{quote}

% \begin{quote}
%     {\tt \textbackslash{}section*\{Contribution Statement\}}
% \end{quote}

\section*{Contribution Statement}
Manqing Dong played a pivotal role in drafting, designing, and deploying the proposed parameter tuning optimization targets and the general automation framework. Additionally, she conducted experiments for AE and Informer using public datasets. Zhanxiang Zhao conducted the main experiments, including the evaluation on eBay's production dataset and experiments with MA, EVT, CNN, and LSTM on public datasets. Yitong Geng made contributions to the feature engineering for model selection and conducted experiments with Elastic Net, Prophet, and LOF on the public datasets. Wentao Li contributed to the experiments with DBSCAN on the public dataset, and provided valuable suggestions for the logical flow of the paper, as well as verifying the business and customer needs both on paper and in real-life scenarios. Wei Wang significantly contributed to the development of the user interface and backend engineering for the fine-tuning function. Huai Jiang provided numerous valuable suggestions regarding the overall idea for the automl service and the organization of the paper.

\section*{Acknowledgments}
We would like to acknowledge the contributions of Huibin Duan for his work on model management and job management, Yuting Tan for her contributions to the data dumper on our online platform, and Yuan Li for her substantial contribution to the user interface design of our anomaly detection platform. Their efforts and expertise have greatly contributed to the success of this research project.

% \appendix
% \section{Appendix}

\color{black}

\bibliographystyle{named}
\bibliography{ijcai23}

\end{document}